%% file: main_arxiv.tex
\documentclass[runningheads]{llncs}
\usepackage{graphicx}

\usepackage[accsupp]{axessibility}  
\usepackage[pagebackref,breaklinks,colorlinks]{hyperref}

\usepackage[width=122mm,left=12mm,paperwidth=146mm,height=193mm,top=12mm,paperheight=217mm]{geometry}

\usepackage{tikz}
\input{package}
\input{macros}
\usepackage[capitalize]{cleveref}
\crefname{section}{Sec.}{Secs.}
\Crefname{section}{Section}{Sections}
\Crefname{table}{Table}{Tables}
\crefname{table}{Tab.}{Tabs.}

\newcommand\blfootnote[1]{%
	\begingroup
	\renewcommand\thefootnote{}\footnote{#1}%
	\addtocounter{footnote}{-1}%
	\endgroup
}

\begin{document}
\pagestyle{headings}
\mainmatter

\title{AutoAvatar: Autoregressive Neural Fields \\for Dynamic Avatar Modeling}

\titlerunning{AutoAvatar: Autoregressive Neural Fields \\for Dynamic Avatar Modeling}

\author{Ziqian Bai$^{1,2*}$ \quad
Timur Bagautdinov$^{2}$ \quad
Javier Romero$^{2}$ \quad
Michael Zollhöfer$^{2}$ \\
Ping Tan$^{1}$ \quad
Shunsuke Saito$^{2}$}

\institute{$^{1}$Simon Fraser University \qquad
$^{2}$Reality Labs Research}

\authorrunning{Z. Bai et al.}

\maketitle

\input{sections/abstract}

\input{sections/introduction}
\input{sections/related}
\input{sections/method}
\input{sections/result}
\input{sections/conclusion}

\clearpage

\bibliographystyle{splncs04}
\bibliography{egbib}

\clearpage

\appendix
{\noindent\Large\textbf{Appendix}}

\input{supplemental/supp_content}
\end{document}

%% file: package.tex
\usepackage{color}
\usepackage{ifthen}
\usepackage{float}
\usepackage{stfloats}
\usepackage{alltt}
\usepackage{newlfont} 
\usepackage{wrapfig}
\usepackage{booktabs}
\usepackage{multirow}
\usepackage{comment}
\usepackage{gensymb}

\usepackage{balance}
\usepackage{comment}
\usepackage{amsmath}
\usepackage{amssymb}
\usepackage{subcaption,siunitx,booktabs}
\usepackage{times}
\usepackage{epsfig}
\usepackage{graphicx}
\usepackage{amsmath}
\usepackage{amssymb}
\usepackage{subcaption}
\usepackage{microtype}
\usepackage{blindtext}
\usepackage{array}
\usepackage{colortbl}
\usepackage{bm}

\usepackage{lipsum}

\usepackage{enumitem}

%% file: macros.tex
\definecolor{DeltaColor}{rgb}{0.039,0.73,0.71}
\definecolor{SetaColor}{rgb}{0.867, 0.0235, 0.376}
\definecolor{SigmaColor}{rgb}{0.98,0.45,0.0}
\definecolor{HaoColor}{rgb}{0.8,0,0}
\definecolor{AlphaColor}{rgb}{0,0,0.8}
\definecolor{BetaColor}{rgb}{0.8,0,0.8}
\definecolor{GammaColor}{rgb}{0.5,0,0.7}
\definecolor{EpsilonColor}{rgb}{0.353,0.725,0.906}
\definecolor{TauColor}{rgb}{0.423,0.235,0.192}
\definecolor{PsiColor}{rgb}{0.8,0.4,0.4}
\newcommand{\zb}[1]{{}}
\newcommand{\tb}[1]{{}}
\newcommand{\jr}[1]{{}}
\newcommand{\sh}[1]{{}}
\newcommand{\pt}[1]{{}}
\newcommand{\MZ}[1]{{}}
\newcommand{\nothing}[1]{}

\definecolor{AudioColor}{rgb}{0.56,0.34,0.62}

\definecolor{DeadlineColor}{rgb}{0.9,0.4,0} 

\definecolor{figred}{rgb}{1,0,0}
\definecolor{figgreen}{rgb}{0,0.6,0}
\definecolor{figblue}{rgb}{0,0,1}
\definecolor{figpink}{rgb}{1,0.63,0.63}

\newcounter{pccount}
\floatstyle{plain}

\newcommand{\filename}[1]{\url{#1}}
\newcommand{\foldername}[1]{\url{#1}}

\newcommand{\partitle}[1]{{\noindent}{\textbf{#1}}.}

%% file: sections/abstract.tex
\begin{abstract}
Neural fields such as implicit surfaces have recently enabled avatar modeling from raw scans without explicit temporal correspondences.
In this work, we exploit autoregressive modeling to further extend this notion to capture dynamic effects, such as
soft-tissue deformations.
Although autoregressive models are naturally capable of handling dynamics, it is non-trivial to apply them to implicit representations, 
as explicit state decoding is infeasible due to prohibitive memory requirements.
In this work, for the first time, we enable autoregressive modeling of implicit avatars.
To reduce the memory bottleneck and efficiently model dynamic implicit surfaces, we introduce the notion 
of articulated observer points, which relate implicit states to the explicit surface of a parametric human body model.
We demonstrate that encoding implicit surfaces as a set of height fields defined on articulated observer points
leads to significantly better generalization compared to a latent representation.
The experiments show that our approach outperforms the state of the art, achieving plausible dynamic deformations even for unseen motions.
\small{\url{https://zqbai-jeremy.github.io/autoavatar}}.

\end{abstract}

\blfootnote{$^{*}$Work done while Ziqian Bai was an intern at Reality Labs Research, Pittsburgh, PA, USA.}

%% file: sections/introduction.tex
\section{Introduction}
\label{sec:intro}
Animatable 3D human body models are key enablers for various applications ranging from virtual try-on to social telepresence~\cite{bagautdinov2021driving}.
While modeling of human avatars from 3D scans without surface registration is gaining more and more attention in recent years~\cite{saito2021scanimate,ma2021scale,tiwari2021neural,chen2021snarf,ma2021pop}, complex temporal dynamics are often 
completely ignored and the resulting deformations are often treated exclusively as a function of the pose parameters.
However, the body shape is not uniquely determined by the current pose of the human, but also depends on the history of 
shape deformations due to secondary motion effects.
The goal of our work is to realistically model these history-dependent dynamic effects for human bodies without requiring 
precise surface registration.

\input{fig/teaser}

To this end, we propose AutoAvatar, a novel autoregressive model for dynamically deforming human bodies.
AutoAvatar models body geometry implicitly - using a signed distance field (SDF) - and
is able to directly learn from raw scans without requiring temporal correspondences 
for supervision.
In addition, akin to physics-based simulation, AutoAvatar infers the complete shape of an avatar given 
history of shape and motion.
The aforementioned properties lead to a generalizable method that models 
complex dynamic effects including inertia and elastic deformations without requiring a 
personalized template or precise temporal correspondences across training frames.

To model temporal dependencies in the data, prior work has typically resorted to autoregressive models~\cite{pons2015dyna,loper2015smpl,martinez2017human,santesteban2020softsmpl}.
While the autoregressive framework naturally allows for incorporation of 
temporal information, combining it with neural implicit surface 
representations~\cite{mescheder2019occupancy,park2019deepsdf,chen2018implicit_decoder} 
for modeling human bodies is non-trivial.
Unlike explicit shape representations, such neural representations implicitly encode the shape 
in the parameters of the neural network and latent codes. 
Thus, in practice, producing the actual shape requires expensive neural network evaluation
at each voxel of a dense spatial grid~\cite{park2019deepsdf}.
This aspect is particularly problematic for autoregressive modeling, since most of the successful 
autoregressive models rely on rollout training~\cite{martinez2017human,ling2020character} to
ensure stability of both training and inference.
Unfortunately, rollout training requires multiple evaluations of the model for each time step, and
thus becomes prohibitively expensive both in terms of memory and compute as the resolution of 
the spatial grid grows.
Another approach would be learning an autoregressive model using latent embeddings that encode 
dynamic shape information~\cite{humanMotionKanazawa19}.
However, it is infeasible to observe the entire span of possible surface deformations from limited real-world scans, 
which makes the model prone to overfitting and leads to worse generalization at test time.

By addressing these limitations, we, for the first time, enable autoregressive training of a full-body geometry model 
represented by a neural implicit surface.
To tackle the scalability issues of rollout training for implicit representations, we introduce the novel notion of articulated observer points.
Intuitively, articulated observer points are temporally coherent locations on the human body surface which store
the dynamically changing state of the implicit function.
In practice, we parameterize the observer points using the underlying body
model~\cite{loper2015smpl}, and then represent the state of the implicit surface as signed heights 
with respect to the vertices of the pose-dependent geometry produced by the articulated model (see \cref{fig:ISE}).
The number of query points is significantly lower than the number of voxels 
in a high-resolution grid, which allows for a significant reduction in terms of memory
and compute requirements, making rollout training tractable for implicit surfaces.
In addition, we demonstrate that explicitly encoding shapes as signed height fields is less prone to overfitting 
compared to latent embeddings, a common way to represent autoregressive states~\cite{ling2020character,xiang2021modeling}. 

Our main contributions are the following:
\begin{itemize}
    \item The first autoregressive approach for modeling history-dependent implicit surfaces of human bodies,
    \item Articulated observer points to enable autogressive training of neural fields, and 
    \item Extensive experiments showing that our approach outperforms existing methods both on shape interpolation and extrapolation tasks.
\end{itemize}

%% file: fig/teaser.tex
\begin{figure*}[!t]
	\centering
    \includegraphics[width=0.95\linewidth]{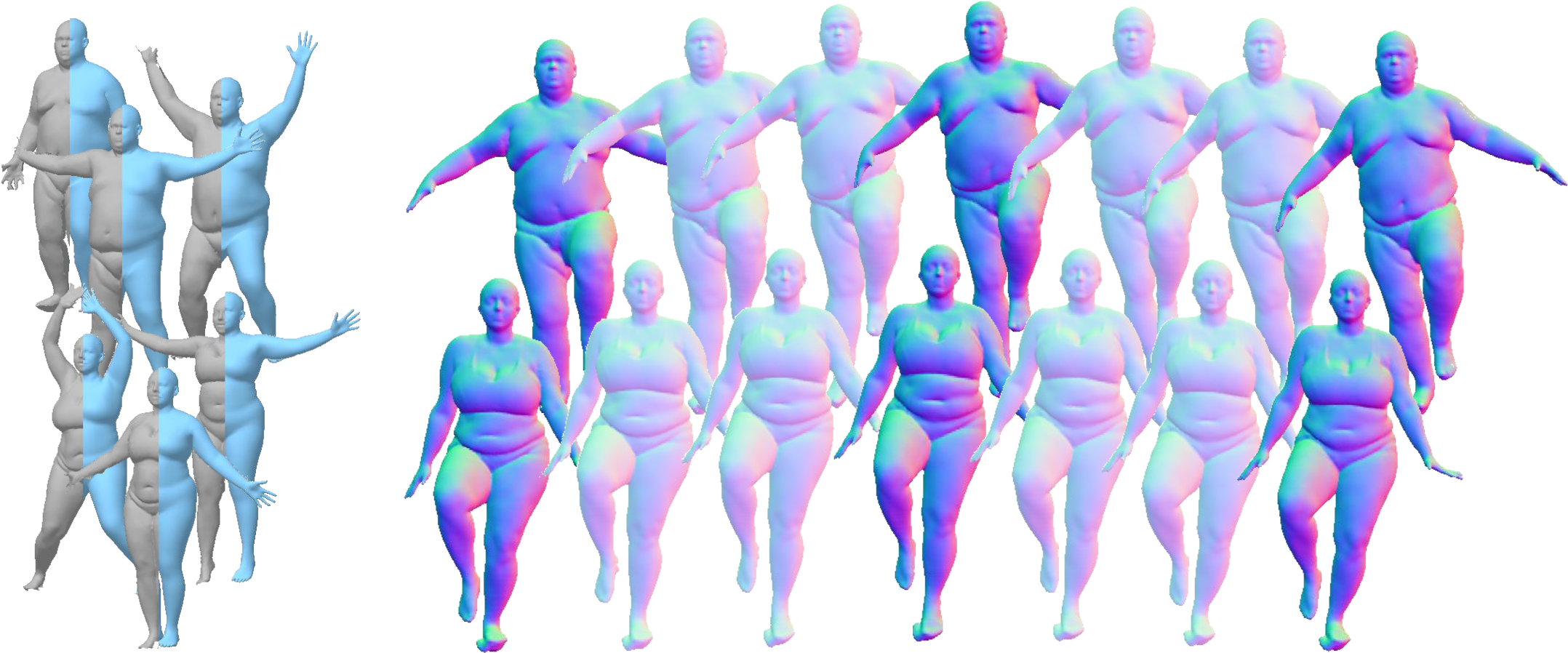}
    \caption{\small{\textbf{AutoAvatar.} Given raw 4D scans with self-intersections, holes, and noise (grey meshes) and fitted SMPL models (blue meshes), AutoAvatar automatically learns highly detailed animatable body models with plausible secondary motion dynamics without requiring a personalized template or surface registration (right).}}
	\label{fig:teaser}
    \vspace{-12pt}
\end{figure*}

%% file: sections/related.tex
\section{Related Work}
\label{sec:related}

\paragraph{Parametric Human Models}

Since the anatomical structure of humans is shared across identities, various methods have been proposed to parameterize shape and pose of human bodies from large-scale 3D scan data~\cite{anguelov2005scape,hasler2009statistical,loper2015smpl,osman2020star,xu2020ghum,alldieck2021imghum}. SCAPE~\cite{anguelov2005scape,hasler2009statistical} learns statistical human model models using triangle deformations. The pioneering work by Allen et al.~\cite{allen2006} used a vertex-based representation enhanced with pose-dependent deformations, but the model was complex and trained with insufficient data, resulting in overfitting. SMPL~\cite{loper2015smpl} improved the generalizability of~\cite{allen2006} by training on more data and removing the shape dependency in the pose-dependent deformations. More recent works show that sparsity in the pose correctives reduces spurious correlations~\cite{osman2020star}, and that non-linear deformation bases parameterized by neural networks achieve better modeling accuracy~\cite{xu2020ghum}.
While most works focus on modeling static human bodies under different poses, Dyna~\cite{pons2015dyna} and DMPL (Dynamic SMPL)~\cite{loper2015smpl} enable parametric modeling of dynamic deformations by learning a linear autoregressive model. Kim et al.~\cite{Kim:2017:VSMPL} combine a volumetric parametric model, VSMPL, with an external layer driven by the finite element method to enable soft tissue dynamics. SoftSMPL~\cite{santesteban2020softsmpl} learns a more powerful recurrent neural network to achieve better generalization to unseen subjects. Xiang et al.~\cite{xiang2021modeling} model dynamically moving clothing from a history of poses. Importantly, the foundation of the aforementioned works is accurate surface registration of a template body mesh~\cite{bogo2014faust,bogo2017dynamic}, which remains non-trivial. Habermann et al.~\cite{habermann2021real} also model dynamic deformations from a history of poses. While they relax the need of registration by leveraging image-based supervision, a personalized template is still required as a preprocessing step.

Recently, neural networks promise to enable the modeling of animatable bodies without requiring surface registration or a personalized template~\cite{deng2019neural,ma2021scale,saito2021scanimate,tiwari2021neural,chen2021snarf}. These methods leverage structured point clouds~\cite{ma2021scale,ma2021pop,zakharkin2021point} or 3D neural fields~\cite{xie2021neural} to learn animatable avatars. Approaches based on neural fields parameterize human bodies as compositional articulated occupancy networks~\cite{deng2019neural} or implicit surface in canonical space with linear blend skinning~\cite{LEAP:CVPR:21,saito2021scanimate,chen2021snarf,wang2021metaavatar} and deformation fields~\cite{tiwari2021neural,palafox2021npms}. Since implicit surfaces do not require surface correspondences for training, avatars can be learned from raw scans. Similarly, neural radiance fields~\cite{mildenhall2020nerf} have been applied to body modeling to build animatable avatars from multi-view images~\cite{peng2021neural,liu2021neural}. However, these approaches represent avatars as a function of only pose parameters, and thus are unable to model dynamics. While our approach is also based on 3D neural fields to eliminate the need for surface registration, our approach learns not only pose-dependent deformations but also history-dependent dynamics by enabling autoregressive training of neural implicit surfaces.

\paragraph{Learning Dynamics}
Traditionally, physics-based simulation~\cite{sifakis2012fem} is used to model dynamics of objects. While material parameters of physics simulation can be estimated from real data~\cite{bhat2003estimating,wang2011data,srinivasan2021learning,yang2017}, accurately simulating dynamic behavior of objects remains an open question. In addition, authenticity of physics-based simulation is bounded by the underlying model, and complex anisotropic materials such as the human body are still challenging to model accurately. For this reason, several works attempt to substitute a deterministic physics-based simulation with a learnable module parameterized by neural networks~\cite{holden2019subspace,zheng2021deepemu,sanchez2020learning,pfaff2020learning}. Such approaches have been applied to cloth simulation~\cite{holden2019subspace,pfaff2020learning}, fluid~\cite{sanchez2020learning}, and elastic bodies~\cite{zheng2021deepemu}. Subspace Neural Physics~\cite{holden2019subspace} learns a recurrent neural network from offline simulation to predict the simulation state in a subspace. Deep Emulator~\cite{zheng2021deepemu} first learns an autoregressive model to predict deformations using a simple primitive (sphere), and applies the learned function to more complex characters. While we share the same spirit with the aforementioned works by learning dynamic deformations in an autoregressive manner, our approach fundamentally differs from them. The aforementioned approaches all assume that physical quantities such as vertex positions are observable with perfect correspondence in time, and thus results are only demonstrated on synthetic data. In contrast, we learn dynamic deformations from real-world observation while requiring only coarse temporal guidance by the fitted SMPL models.
This property is essential to model faithful dynamics of real humans.

%% file: sections/method.tex
\section{Method}
\label{sec:method}
\input{fig/overview}

Our approach is an autoregressive model, which takes as inputs human poses and a shape history 
and produces the implicit surface for a future frame.
\cref{fig:overview} shows the overview of our approach.
Given a sequence of $T$ implicitly encoded shapes $\{\bm{S}_{t-T+1}, ..., \bm{S}_{t}\}$ and 
$T+1$ poses $\{\bm{p}_{t-T+1}, ..., \bm{p}_{t+1}\}$ with $t$ being the current time frame, 
our model predicts the implicit surface $\bm{S}_{t+1}$ of the future frame $t+1$.
The output shape $\bm{S}_{t+1}$ is then passed as an input to the next frame prediction in an autoregressive manner. 
Our model is supervised directly with raw body scans, and requires a training dataset of 4D
scans (sequences of 3D scans) along with fitted SMPL body models~\cite{loper2015smpl}.
Unfortunately, explicitly representing shapes $\bm{S}_t$ as levelsets of implicit surface is 
prohibitively expensive for end-to-end training. 
To this end, we introduce the concept of \textit{articulated observer points} - vertex 
locations on the underlying articulated model - which are used as a local reference for 
defining the full body geometry.
The underlying implicit surface is encoded as a height field with respect to the 
articulated observer points (\cref{sec:imp_enc}). 
Given a history of height fields and pose parameters, we convert those to dynamic
latent feature maps in UV space (\cref{sec:dyn_inf}). 
Finally, we map the resulting features to SDFs by associating continuous 3D space with the learned features on the SMPL vertices, which are directly supervised by point clouds with surface normals (\cref{sec:imp_dec}).

\input{fig/method}

\input{sections/03_1_shape_encoding}
\input{sections/03_2_dynamic_feat}
\input{sections/03_3_shape_decode}
\input{sections/03_4_training}

%% file: fig/overview.tex
\begin{figure*}[!t]
	\centering
    \includegraphics[width=\linewidth]{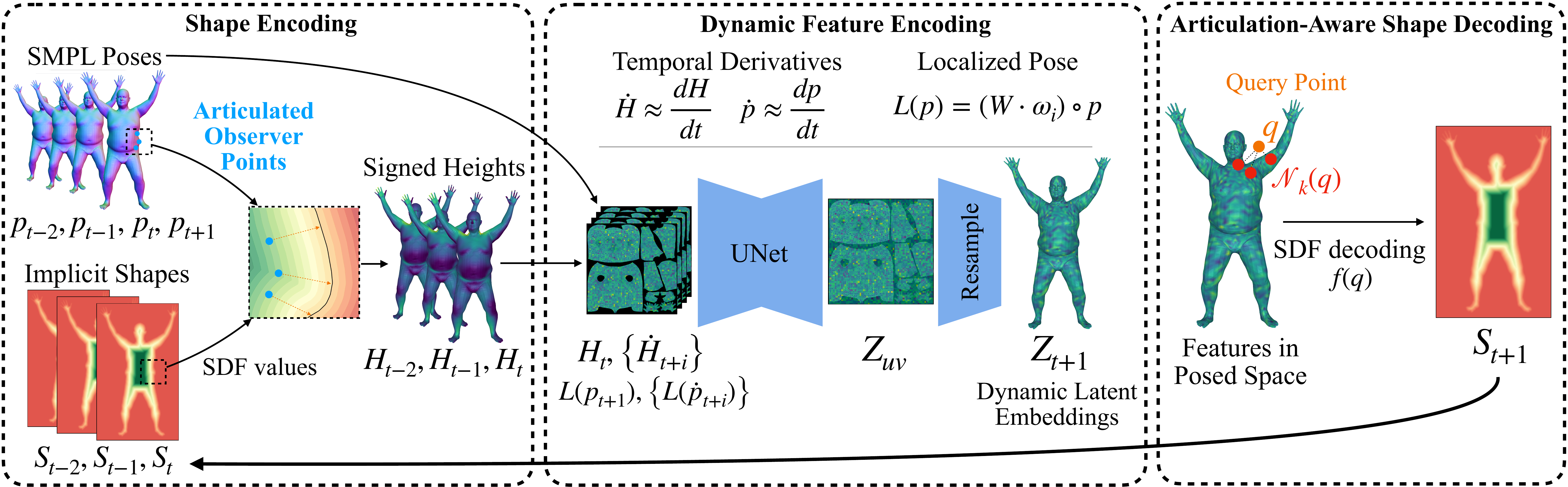}
    \caption{\small{\textbf{Overview.} AutoAvatar learns a pose-driven animatable human body model with plausible dynamics including secondary motions. Notice that our approach takes the history of implicit shapes in an autoregressive manner for learning dynamics.
    }}
	\label{fig:overview}
    \vspace{-12pt}
\end{figure*}

%% file: fig/method.tex
\begin{figure}[!t]
	\centering
    \begin{subfigure}{.35\textwidth}
      \centering
      \includegraphics[height=3.4cm]{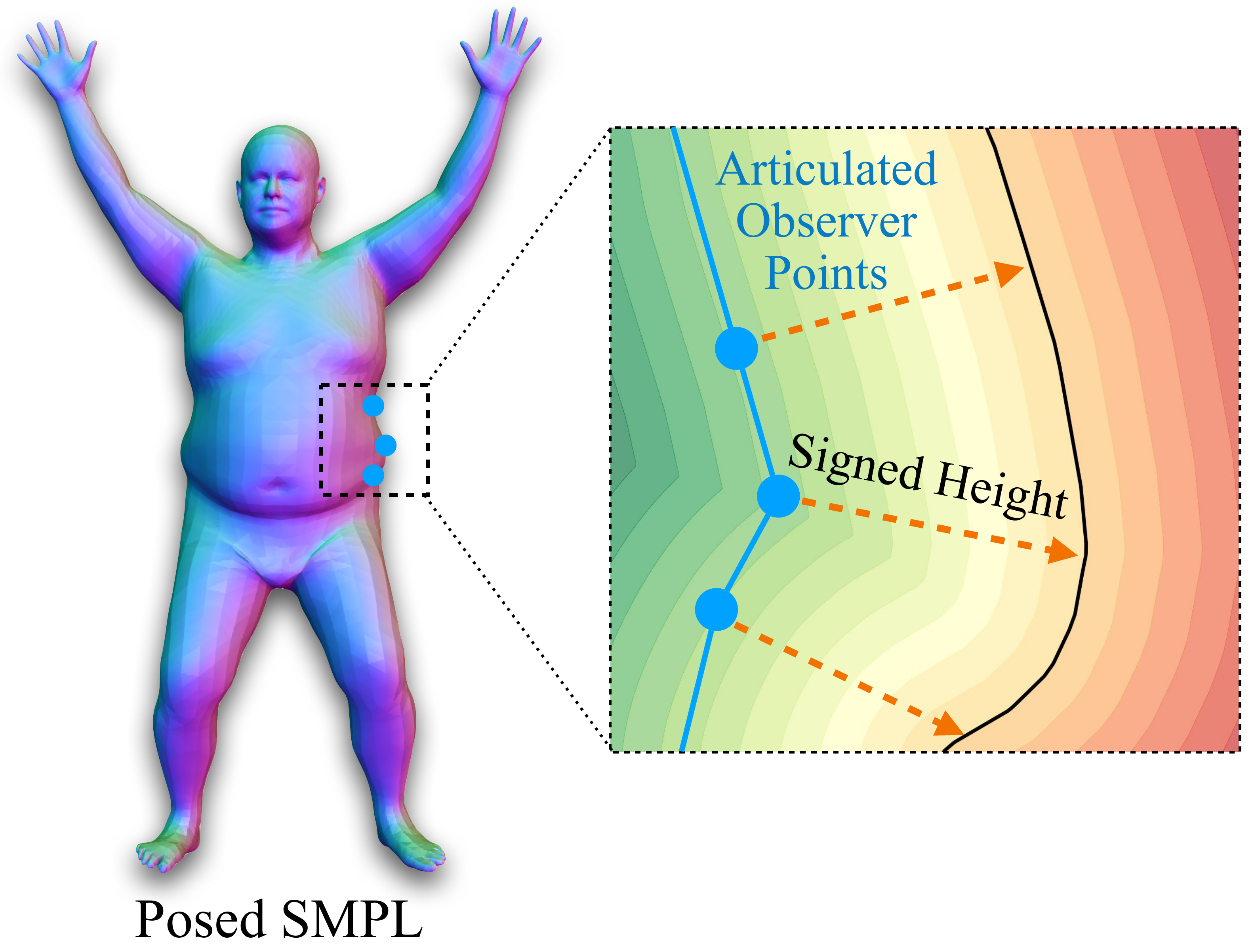}
      \caption{\small{Shape Encoding.}}
      \label{fig:ISE}
    \end{subfigure}%
    \begin{subfigure}{.65\textwidth}
      \centering
      \includegraphics[height=3.4cm]{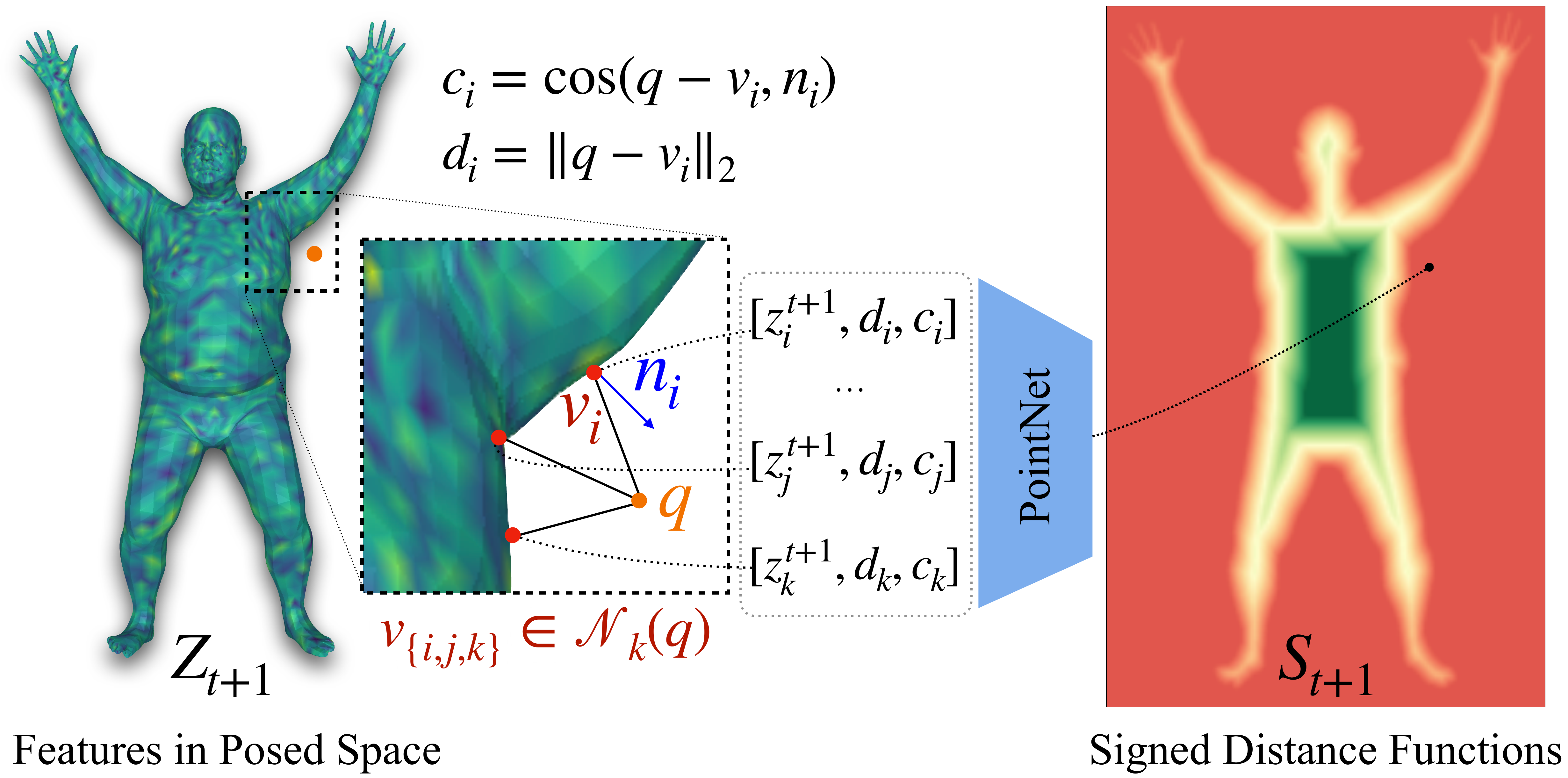}
      \caption{\small{Articulation-Aware SDF Decoding.}}
      \label{fig:F2I}
    \end{subfigure}
    \caption{\small{\textbf{Shape Encoding/Decoding.} Our novel shape encoding via articulated observer points and articulated-aware SDF decoding lead to faithful modeling of dynamics.}}
	\label{fig:IFC}
    \vspace{-12pt}
\end{figure}

%% file: sections/03_1_shape_encoding.tex
\subsection{Shape Encoding via Articulated Observer Points}
\label{sec:imp_enc}
The core of our approach is an autoregressive model that operates
on implicit neural surfaces, allowing us to incorporate temporal shape information 
necessary for modeling challenging dynamics.
The key challenge that arises when training such autoregressive models is finding a 
way to encode the shape - parameterized implicitly as a neural field - into a representation 
that can be efficiently computed and fed back into the model.
The most straightforward way is to extract an explicit geometry 
representation by evaluating the neural field on a dense spatial grid and 
running marching cubes~\cite{lorensen1987marching}.
However, in practice this approach is infeasible due to prohibitive memory and computational 
costs, in particular due to the cubic scaling with respect to the grid dimensions. 
Instead, we propose to encode the state of the implicit surface into a set of observer points.

Encoding geometry into discrete point sets has been shown to be efficient and
effective for learning shape representations from point clouds~\cite{prokudin2019efficient}. 
Prokudin et al.~\cite{prokudin2019efficient} relies on a fixed set of randomly sampled observer points in global world coordinates, which is not suitable for modeling dynamic humans due to the articulated nature of human motion.
Namely, a model relying on observer points with a fixed 3D location needs 
to account for extremely large shape variations including rigid transformations, 
making the learning task difficult. 
Moreover, associating randomly sampled 3D points with a parametric human body is non-trivial.
To address these limitations, we further extend the notion of observer points to an articulated
template represented by the SMPL model~\cite{loper2015smpl}, which provides several advantages for modeling dynamic articulated geometries.
In particular, soft-tissue dynamic deformations appear only around the minimally clothed body, 
and we can rely on this notion as an explicit prior to effectively allocate
observer points only to the relevant regions. 
In addition, the SMPL model provides a mapping of 3D vertices to a common UV parameterization, 
allowing us to effectively process shape information using 2D CNNs in a temporally consistent manner.

More specifically, to encode the neural implicit surface into the articulated observer points, 
we compute ``signed heights" $\bm{H} = \{h_i\}_{i=1}^M \in \mathbb{R}^{M}$ from $M$ vertices on a fitted SMPL model. 
For each vertex, the signed height $h_i$ is the signed distance from the vertex 
to the zero-crossing of the implicit surface along the vertex normal (see \cref{fig:ISE}). 
We use the iterative secant method as in~\cite{DVR} to compute the zero-crossings. 
Note that there can be multiple valid signed heights per vertex since the line along the normal
can hit the zero-crossing multiple times. 
Based on the observation that the SMPL vertices are usually close to the actual surface with their normals roughly facing into the same direction, we use the minimum signed height within a predefined range $[h_{min}, h_{max}]$ (in our experiments, we use $h_{min} = -2cm, h_{max} = 8cm$). If no zero-crossing is found inside this range, we set the signed height to $h_{min}$. 
Note that the computed heights are signed because the fitted SMPL can go beyond the actual surface due to its limited expressiveness and inaccuracy in the fitting stage.

%% file: sections/03_2_dynamic_feat.tex
\subsection{Dynamic Feature Encoding}
\label{sec:dyn_inf}

The essence of AutoAvatar is an animatable autoregressive model. 
In other words, a reconstructed avatar is driven by pose parameters, while secondary dynamics 
is automatically synthesized from the history of shapes and poses. 
To enable this, we learn a mapping that encodes the history of shape and pose information
to latent embeddings containing the shape information of the future frame.
More specifically, denoting the current time frame as $t$, we take as input $T+1$ poses $\{\bm{p}_{t-T+1}, ..., \bm{p}_{t+1}\}$ and $T$ signed heights vectors $\{\bm{H}_{t-T+1}, ..., \bm{H}_{t}\}$, and produce dynamic features $\bm{Z}_{t+1} \in \mathbb{R}^{M \times C}$.
Given these inputs, we also compute the temporal derivatives of poses $\{\dot{\bm{p}}_{t+i}\}_{i=-T+2}^{1}$ and signed heights $\{\dot{\bm{H}}_{t+i}\}_{i=-T+2}^0$ as follows:
\begin{equation}
\begin{aligned}
    \dot{\bm{p}}_{k} &= \bm{p}_{k} \bm{p}_{k-1}^{-1} \\
    \dot{\bm{H}}_{k} &= \bm{H}_{k} - \bm{H}_{k-1}.
\end{aligned}
\end{equation}
Note that in practice $\bm{p}_{*}$ are represented as quaternions, and $\bm{p}_{k} \bm{p}_{k-1}^{-1}$ is computed by first converting multipliers to rotation matrices, multiplying those, and then converting the 
product back to quaternions.
To emphasize small values in $\dot{\bm{H}}$, we apply the following transformation 
$g(x) = \text{sign}(x) \cdot \text{ln}(\alpha |x| + 1) \cdot \beta$, 
where $\alpha = 1000$ and $\beta = 0.25$. 
Following prior works~\cite{saito2021scanimate,bagautdinov2021driving}, we also localize pose
parameters to reduce long range spurious correlations as follows:
\begin{equation}
\begin{aligned}
    L(\bm{p}) &= (W \cdot \omega_i) \circ \bm{p},
\end{aligned}
\end{equation}
where $\circ$ denotes the element-wise product, $i$ is the vertex index, $W \in \mathbb{R}^{J \times J}$ is an association matrix of $J$ joints, and $\omega_i \in \mathbb{R}^{J \times 1}$ is the skinning weights of the $i$-th vertex. 
We set $W_{n,m} = 1$ if the $n$-th joint is within the $1$-ring neighborhood of the $m$-th joint (otherwise $W_{n,m} = 0$). 
Note that the derivative of the root transformation is included in $\{L(\dot{\bm{p}}_{t+i})\}$ without localization. 
Finally, we map $\bm{H}_{t}$, $\{\dot{\bm{H}}_{t+i}\}$, $L(\bm{p}_{t+1})$, and 
$\{L(\dot{\bm{p}}_{t+i})\}$ to UV space using barycentric interpolation. 
The concatenated features are fed into a UNet~\cite{ronneberger2015u} to generate a 
feature map $\bm{Z}_{uv}$. 
We then resample $\bm{Z}_{uv}$ on the UV coordinates corresponding to SMPL vertices 
to obtain the per-vertex dynamic latent embeddings $\bm{Z}$. 
We empirically found that incorporating temporal derivatives further improves the realism
of dynamics (see Supp. Mat. video for comparison). 

%% file: sections/03_3_shape_decode.tex
\subsection{Articulation-Aware Shape Decoding}
\label{sec:imp_dec}

Given the dynamic feature $\bm{Z}_{t+1}=\{\bm{z}^{t+1}_1,...,\bm{z}^{t+1}_M\}$ and a query point $\bm{q}$, we decode signed distance fields $f(\bm{q})$ to obtain the surface geometry of the dynamic avatar. 
Several methods model the implicit surface in canonical space by jointly learning a warping function from the posed space to the canonical space~\cite{saito2021scanimate,chen2021snarf}. 
However, we observe that the canonicalization step is very sensitive to small fitting error in the SMPL model, and further amplifies the error in the canonical space, making it difficult to learn dynamics (see discussion in \cref{sec:result}). 
Therefore, we directly model the implicit surface in a posed space while being robust to pose changes.
Inspired by Neural Actor~\cite{liu2021neural}, we associate a queried 3D point with a human body model and pose-agnostic spatial information.
Specifically, Neural Actor uses height from the closest surface point on the SMPL model to the query location together with a feature vector sampled on the same surface point. 
However, we find that their approach based on the single closest point leads to artifacts around body joints (e.g., armpits) for unseen poses.
To better distinguish regions with multiple body parts, we instead use $k$-nearest neighbor vertices. \cref{fig:F2I} shows the illustration of our SDF decoding approach.
Given a query point $\bm{q}$, we first compute the k-nearest SMPL vertices $\{\bm{v}_i\}_{i \in \mathcal{N}_k(\bm{q})}$, where $\mathcal{N}_k(\bm{q})$ is a set of indices of k-nearest neighbor vertices. 
To encode pose-agnostic spatial information, we use rotation-invariant features. Specifically,
we compute the distance $d_i=\|\bm{q}-\bm{v}_i\|_2$ and cosine value $c_i=\mathrm{cos}(\bm{x}_i, \bm{n}_i)$, where $\bm{x}_i$ is the vertex-to-query vector $\bm{x}_i = \bm{q} - \bm{v}_i$, and $\bm{n}_i$ is the surface normal on $\bm{v}_i$. 
We feed the concatenated vector $\left[\bm{z}^{t+1}_i, d_i, c_i\right]$ into a PointNet-like~\cite{qi2017pointnet} architecture to compute the final SDFs with the max pooling replaced by a weighted average pooling based on jointly predicted weights for better continuity.

As in~\cite{saito2021scanimate}, we employ implicit geometric regularization (IGR)~\cite{gropp2020implicit} to train our model directly from raw scans without requiring watertight meshes. Note that in contrast, other methods~\cite{LEAP:CVPR:21,chen2021snarf,tiwari2021neural} require watertight meshes to compute ground-truth occupancy or signed distance values for training. Our final objective function $L$ is the following:
\begin{equation}
\label{eq:igr}
\begin{aligned}
    L &= L_{s} + L_{n} + \lambda_{igr} L_{igr} + \lambda_{o} L_{o}, \\
\end{aligned}
\end{equation}
where  $\lambda_{igr} = 1.0$, $\lambda_{o} = 0.1$.
$L_{s}$ promotes SDFs which vanish on the ground truth surface, while $L_{n}$ encourages that its normal align with the ones from data:
$L_{s} = \sum_{\bm{q} \in \bm{Q}_{s}} |f(\bm{q})|$, $L_{n} = \sum_{\bm{q} \in \bm{Q}_{s}} \|\nabla_{\bm{q}} f(\bm{q}) - \bm{n}(\bm{q})\|_2$,
where $\bm{Q}_{s}$ is the surface of the input raw scans. $L_{igr}$ is the Eikonal regularization term~\cite{gropp2020implicit} that encourages the function $f$ to satisfy the Eikonal equation: $L_{igr} = \mathbb{E}_{\bm{q}}\,(\|\nabla_{\bm{q}} f(\bm{q})\|_2 - 1)^2$, and $L_o$ prevents off-surface SDF values from being too close to the zero-crossings as follows: $L_{o} = \mathbb{E}_{\bm{q}}\,(\text{exp}(-\gamma \cdot |f(\bm{q})|))$, where $\gamma = 50$.

%% file: sections/03_4_training.tex
\subsection{Implementation Details}
\label{sec:training}

\noindent\textbf{Network Architectures.}
In our experiments, we use a UV map of resolution $256\times256$, $T=3$, and $k=20$. To reduce the imbalance of SMPL vertex density for k-NN computation, we use $3928$ points subsampled by poisson-disk sampling on the SMPL mesh. Before being fed into the UNet, $L(\bm{p}_{k+1})$ and $\{L(\dot{\bm{p}}_{t+i})\}$ are compressed to $32$ channels using $1\times1$ convolutions. The UNet uses convolution and transposed convolution layers with untied biases, a kernel size of $3$, no normalization, and LeakyReLU with a slope of $0.2$ as the non-linear activation, except for the last layer which uses TanH. The SDF decoder is implemented as an MLP, which takes as input $64$-dim features from the UNet, positional encoded $d_i$ and $c_i$ up to 4-th order Fourier features. The number of intermediate neurons in the first part of the MLP is $(128, 128, 129)$, where the output is split into a $128$-dim feature vector and a $1$-dim scalar, which is converted into non-negative weights by softmax across the k-NN samples. After weighted average pooling, the aggregated feature is fed into another MLP with a neuron size of $(128, 128, 1)$ to predict the SDF values. The MLPs use Softplus with $\beta=100$ and a threshold of $20$ as non-linear activation except for the last layer which does not apply any activation.

\noindent\textbf{Training.} Our training consists of two stages. First, we train our model using ground-truth signed heights without rollout for $90000$ iterations. Then, we finetune the model using a rollout of 2 frames for another $7500$ iterations to reduce error accumulation for both training and inference. We use the Adam optimizer with a learning rate of $1.0 \times 10^{-4}$ ($1.0 \times 10^{-5}$) at the first (second) stage. To compute $L_n$, we sample $10000 (1000)$ points on the scan surface. Similarly, for $L_{igr}$, we sample $10000 (1000)$ points around the scan surface by adding Gaussian noise with standard deviation of $10$cm to uniformly sampled surface points, and sample $2000 (500)$ points within the bounding box around the raw scans. The points uniformly sampled inside the bounding box are also used to compute $L_o$. Both stages are trained with a batch size of $1$.

\noindent\textbf{Inference.}
At the beginning of the animations, we assume ground-truth raw scans are available for the previous $T$ frames for initialization. If no ground truth initial shape is available, 
we initialize the first $T$ frames with our baseline model conditioned only on pose parameters. Note that the scan data is extremely noisy around the hand and foot areas, and the SMPL fitting of the head region is especially inaccurate. Therefore, we fix the dynamic features on the face, hands, and feet to the ones of the first frame.

%% file: sections/result.tex
\section{Experimental Results}
\label{sec:result}

\subsection{Datasets and Metrics}
\label{sec:datasets_metrics}

\partitle{Datasets}
We use the DFaust dataset~\cite{bogo2017dynamic} for both training and quantitative evaluation, and AIST++~\cite{tsuchida2019aist,li2021ai} for qualitative evaluation on unseen motions.
For the DFaust dataset, we choose 2 subjects ($50002$ and $50004$), who exhibit the most soft-tissue deformations.
The interpolation test evaluates the fidelity of dynamics under the same type of motions as in training but at different time instance,
and the extrapolation test evaluates performance on unseen motion.
For 50002, we use the 2nd half of \texttt{chicken\_wings} and \texttt{running\_on\_spot} for the interpolation test, \texttt{one\_leg\_jump} for the extrapolation test, and the rest for training.
For 50004, we use the 2nd half of \texttt{chicken\_wings} and \texttt{running\_on\_spot} for interpolation, \texttt{one\_leg\_loose} for extrapolation, and the rest for training.
The fitted SMPL parameters in DFaust are provided by the AMASS~\cite{mahmood2019amass} dataset that uses sparse points on the registered data as approximated mocap marker locations and computes the parameters using MoSh~\cite{loper2014mosh}. Note that more accurate pose can be obtained by using all the registration vertices (see \cref{sec:better_poses}), but this is not required by our method to recover soft-tissue deformation.

\partitle{Metrics}
For evaluation, we extract the $0$-level set surface at each time step using Marching Cubes~\cite{lorensen1987marching} with a resolution of $256^3$. We also use simplified scans with
around $10000$ vertices and outlier points (distance to the nearest SMPL vertex larger than $10$cm) have been removed.
We evaluate the accuracy of the predicted surface in terms of its position and dynamics accuracy.
The surface position accuracy is measured by averaging the distance from each simplified scan vertex to the closest prediction surface point.
Evaluating the dynamics accuracy of the implicit surface efficiently is more challenging.
We approximate the local occupied volume as a scalar per registration vertex representing the ratio of surrounding points contained in the interior of the (ground-truth or inferred) surface. We use $10$ points uniformly sampled inside a $5$cm cube centered at the vertex. The head, hands and feet vertices are ignored due to their high noise levels.
The temporal difference of this scalar across adjacent frames can be interpreted as a dynamic measure of the local volume evolution. We report the mean square difference between this dynamic descriptor as computed with the ground truth simplified scan and the inferred implicit surface.
Since a small phase shift in dynamics may lead to large cumulative error, reporting only the averaged errors from the entire frames can be misleading. Therefore, we report errors along the progression of rollout predictions. 
For each evaluation sequence, we start prediction every $20$ frames (i.e., $20$th frame, $40$th frame, ...), and use the ground truth pose and shape history only for the first frame, followed by the autoregressive predictions for the error computation.
In \cref{tab:quant_eval}, we report the averaged errors for both metrics after 1, 2, 4, 8, 16, and 30 rollouts. The errors for small number of rollouts evaluate the accuracy of future shape prediction given the ground-truth shape history, whereas the errors with longer rollouts evaluate the accumulated errors by autoregressively taking as input the predictions of previous frames. We discuss the limitation of error metrics with longer rollouts in \cref{sec:eval}.

\subsection{Evaluation}
\label{sec:eval}

In this section, we provide comprehensive analysis to validate our design choices and highlight the limitations of alternative approaches and SoTA methods based on both implicit and explicit shape representations. Note that all approaches use the same training set, and are trained with the same number of iterations as our method for fair comparison. 

\input{tables/quant_eval}

\noindent\textbf{Effectiveness of Autoregressive Modeling.}
While autoregressive modeling is a widely used technique for learning dynamics~\cite{pons2015dyna,zheng2021deepemu,pfaff2020learning}, 
several recent methods still employ only the history of poses for modeling dynamic 
avatars~\cite{xiang2021modeling,habermann2021real}.
Thus, to evaluate the effectiveness of autoregressive modeling we compare 
AutoAvatar with pose-dependent alternatives that use neural implicit surfaces. 
More specifically, we design the following $3$ non-autoregressive baselines:

\begin{enumerate}[nosep]
  \item Pose: We only feed pose parameters of the next frame $L(\bm{p}_{t+1})$ in our architecture. Prior avatar modeling methods based on neural fields employ this pose-only parameterization~\cite{saito2021scanimate,tiwari2021neural,chen2021snarf}.
  \item PoseTCN: Temporal convolutional networks (TCN)~\cite{lea2016temporal} support the incorporation of a long-range history for learning tasks, and have been used in several avatar modeling methods~\cite{xiang2021modeling,habermann2021real}. Thus, we use a TCN that takes as input the sequence of poses with the length of $16$. We first compute localized pose parameters, as in our method, for each frame and apply the TCN to obtain $64$-dim features for each SMPL vertex. The features are then fed into the UNet and SDF decoders identical to our method. 
  \item Pose+dPose: Our approach without autoregressive components ($\bm{H}_{t}$, $\{\dot{\bm{H}}_{t+i}\}$).
\end{enumerate}

\MZ{It might be good to make even more clear that while the other approaches are better in terms of numbers for larger roll-out, they do not show dynamics. Even in the captions of the tables etc. This is because of a collapse to predicting the "mean", i.e., it does not show dynamics. I.e., while our approaches produces the most "plausible" dynamics, since the future is not uniquely predictable it might deviate from the actual ground truth. All the other methods do not predict dynamics and collapse to the mean.}

\input{fig/vs_nonauto}

\cref{tab:quant_eval} shows that our approach outperforms the baseline methods for the first $8$ frames for interpolation, and first $16$ frames for extrapolation. 
In particular, there is a significantly large margin for the first $4$-$8$ frames, indicating that our method achieves the most accurate prediction of the future frames given the ground-truth shape history. 
We also observe that the non-autoregressive methods tend to collapse to predicting the ``mean" shape under each pose without faithful dynamics for unseen motions (see \cref{fig:non_autoregr} and Supp. Mat. video).
Since the accumulation of small errors in each frame may lead to large deviations 
from the ground-truth due to high sensitivity to initial conditions in dynamic systems~\cite{ott1990controlling}, for longer rollouts mean predictions without 
any dynamics can produce lower errors than more dynamic predictions. 
In fact, although our method leads to slightly higher errors on longer rollouts, \cref{fig:non_autoregr} clearly shows that our approach produces the most visually 
plausible dynamics on the AIST++ sequences. 
Importantly, we do not observe any instability or explosions in our 
autoregressive model for longer rollouts, as can be seen from the error
behavior shown in \cref{fig:vs_softsmpl}.
We also highly encourage readers to see Supp. Mat. video for qualitative comparison
in animation. 
In summary, our results confirm that autoregressive modeling plays a critical 
role for generalization to unseen motions and improving the realism of dynamics.

\input{fig/vs_latent}

\noindent\textbf{Explicit Shape Encoding vs. Latent Encoding.}
Efficiently encoding the geometry of implicit surfaces is non-trivial.
While our proposed approach encodes the geometry via signed heights on articulated 
observer points, prior approaches have demonstrated shape encoding based on a 
learned latent space~\cite{park2019deepsdf,bagautdinov2021driving}. 
Therefore, we also investigate different encoding methods for autoregressive modeling with the following $2$ baselines:

\begin{enumerate}[nosep]
  \item G-embed: Inspired by DeepSDF~\cite{park2019deepsdf}, we first learn per-frame global embeddings $l_g \in \mathbb{R}^{512}$ with the UNet and SDF decoder by replacing $\bm{H}_{t}$, $\{\dot{\bm{H}}_{t+i}\}$, $\{L(\dot{\bm{p}}_{t+i})\}$ with repeated global embeddings. Then, we train a small MLP with three 512-dim hidden layers using Softplus except for the last layer, taking as input $\bm{p}_{t+1}$, $\{\dot{\bm{p}}_{t+i}\}$, and $3$ embeddings of previous frames to predict the global embedding at time $t+1$. 
  \item L-embed: For modeling mesh-based body avatars, localized embeddings are shown to be effective~\cite{bagautdinov2021driving}. Inspired by this, we also train a model with localized embeddings $l_l \in \mathbb{R}^{16\times64\times64}$. We first learn per-frame local embeddings $l_l$ together with the UNet and SDF decoder by replacing $\bm{H}_{t}$, $\{\dot{\bm{H}}_{t+i}\}$, $\{L(\dot{\bm{p}}_{t+i})\}$ with bilinearly upsampled $l_l$. Then we train another UNet that takes as input $L(\bm{p}_{t+1})$, $\{L(\dot{\bm{p}}_{t+i})\}$, and $3$ embeddings of previous frames to predict the localized embeddings at time $t+1$.
\end{enumerate}

Note that for evaluation, we optimize per-frame embeddings for test sequences using \cref{eq:igr} such that the baseline methods can use the best possible history of embeddings for autoregression. \cref{tab:quant_eval} shows that our method outperforms L-embed in all cases because L-embed becomes unstable for the test sequences. For G-embed, we observe the same trend as for the non-autoregressive baselines: our approach achieves significantly more accurate predictions of the future frames given the ground-truth trajectories (see the errors for 1-4 rollouts), and G-embed tends to predict ``mean" shapes without plausible dynamics. The qualitative comparison in \cref{fig:latent_feat_autoregr} confirms that our approach produces more plausible dynamics. Please refer to Supp. Mat. video for detailed visual comparison in animation. 
We summarize that physically meaningful shape encodings (e.g., signed heights) enable more stable learning of dynamics via autoregression than methods relying 
on latent space.

\input{fig/vs_sota_vis}
\input{fig/vs_sota_plot}

\partitle{Comparison to SoTA Methods.}
We compare our approach with state-of-the-art methods for both implicit surface representations and mesh-based representations. As a method using neural implicit surfaces, we choose SNARF~\cite{chen2021snarf}, which jointly learns a pose-conditioned implicit surface in a canonical T-pose and a forward skinning network for reposing. Similar to ours, SNARF does not require temporal correspondences other than the fitted SMPL models. We use the training code released by the authors using the same training data and the fitted SMPL parameters as in our method. Note that in the DFaust experiment in \cite{chen2021snarf}, SNARF is trained using only the fitted SMPL models to DFaust as ground-truth geometry, which do not contain any dynamic deformations. \cref{tab:quant_eval} shows that our approach significantly outperforms SNARF for any number of rollouts. Interestingly, SNARF can produce dynamic effects for training data by severely overfitting to the pose parameters, but this does not generalize to unseen poses as the learned dynamics is the results of spurious correlations. As mentioned in \cref{sec:imp_dec}, we also observe that the performance of SNARF heavily relies on the accuracy of the SMPL fitting for canonicalization, and any small alignment errors in the underlying SMPL registration deteriorates their test-time performance (see \cref{fig:prior_works}). 
Therefore, this experiment demonstrates not only the importance of autoregressive dynamic avatar modeling, but also the efficacy of our articulation-aware shape decoding approach given the quality of available SMPL fitting for real-world scans. 

We also compare against SoftSMPL~\cite{santesteban2020softsmpl}, a state-of-the-art mesh-based method that learns dynamically deforming human bodies from registered meshes. The authors of SoftSMPL kindly provide their predictions on the sequence of \texttt{one\_leg\_loose} for subject 50004, which is excluded from training for both our method and SoftSMPL for fair comparison. To our surprise, \cref{fig:vs_softsmpl} show that our results are slightly better on both metrics for the majority of frames, although we tackle a significantly harder problem because our approach learns dynamic bodies directly from raw scans, whereas SoftSMPL learns from the carefully registered data. We speculate that the lower error may be mainly attributed to the higher resolution of our geometry using implicit surfaces in contrast to their predictions on the coarse SMPL topology (see \cref{fig:prior_works}). Nevertheless, this result is highly encouraging as our approach achieves comparable performance on dynamics modeling without having to rely on surface registration.

%% file: tables/quant_eval.tex
\begin{table}
\caption{\small{\textbf{Quantitative Comparison with Baseline Methods.} 
Our method produces the most accurate predictions of the future frames given the ground-truth shape history among all baseline methods (see rollout $1$-$4$). 
For longer rollouts, more dynamic predictions lead to higher error than 
less dynamic results due to high sensitivity to initial conditions in dynamic systems~\cite{ott1990controlling} (see discussion in \cref{sec:eval}).
}}
  \label{tab:quant_eval}

    \begin{subtable}{\textwidth}
      
  \centering
  \caption{\small{Mean Scan-to-Prediction Distance (mm)     \label{tab:results_dfaust_sd}
$\bm{\downarrow}$ on DFaust.}}
    \setlength{\tabcolsep}{0.5em}
    \begin{tabular}{l|l|c|c|c|c|c|c}
    \toprule
    \multicolumn{2}{c|}{} & \multicolumn{6}{c}{Rollout (\# of frames)} \\
    \multicolumn{2}{c|}{} & 1 & 2 & 4 & 8 & 16 & 30 \\
    \hline
    \multicolumn{8}{c}{\textit{Interpolation Set}} \\
    \hline
    \multirow{4}{*}{Non-Autoregressive} & SNARF~\cite{chen2021snarf} & 7.428 & 7.372 & 7.337 & 7.476 & 7.530 & 7.656  \\
    
    & Pose & 4.218 & 4.202 & 4.075 & 4.240 & 4.409 & 4.426 \\
    
    & PoseTCN & 4.068 & 4.118 & 4.086 & 4.228 & 4.405 & 4.411 \\
    
    & Pose + dPose & 3.852 & 3.841 & 3.764 & 3.972 & \cellcolor[HTML]{FFDB9B}4.164 & \cellcolor[HTML]{FFDB9B}4.156 \\
    \midrule
    
    \multirow{3}{*}{Autoregressive} & G-embed & 2.932 & 3.006 & 3.131 & \cellcolor[HTML]{FFACAC}3.462 & \cellcolor[HTML]{FFACAC}3.756 & \cellcolor[HTML]{FFACAC}3.793 \\
    
    & L-embed & \cellcolor[HTML]{FFDB9B}1.784 & \cellcolor[HTML]{FFDB9B}2.138 & \cellcolor[HTML]{FFDB9B}2.863 & 4.250 & 5.448 & 5.916 \\
    
    
    & Ours & \cellcolor[HTML]{FFACAC}1.569 & \cellcolor[HTML]{FFACAC}1.914 & \cellcolor[HTML]{FFACAC}2.587 & \cellcolor[HTML]{FFDB9B}3.627 & 4.736 & 5.255 \\
    
    \hline
    \multicolumn{8}{c}{\textit{Extrapolation Set}} \\
    \hline
    \multirow{4}{*}{Non-Autoregressive} & SNARF~\cite{chen2021snarf} & 7.264 & 7.287 & 7.321 & 7.387 & 7.308 & 7.251 \\
    
    & Pose & 4.303 & 4.306 & 4.308 & 4.299 & 4.385 & 4.398 \\
    
    & PoseTCN & 4.090 & 4.091 & 4.105 & 4.119 & 4.233 & 4.257 \\
    
    & Pose + dPose & 3.984 & 3.991 & 4.017 & 4.063 & 4.162 & \cellcolor[HTML]{FFDB9B}4.190 \\
    \midrule
    
    \multirow{3}{*}{Autoregressive} & G-embed & 2.884 & 2.926 & 3.043 & \cellcolor[HTML]{FFDB9B}3.258 & \cellcolor[HTML]{FFACAC}3.577 & \cellcolor[HTML]{FFACAC}3.787 \\
    
    & L-embed & \cellcolor[HTML]{FFDB9B}1.329 & \cellcolor[HTML]{FFDB9B}1.539 & \cellcolor[HTML]{FFDB9B}2.079 & 3.326 & 4.578 & 5.192 \\
    
    
    & Ours & \cellcolor[HTML]{FFACAC}1.150 & \cellcolor[HTML]{FFACAC}1.361 & \cellcolor[HTML]{FFACAC}1.834 & \cellcolor[HTML]{FFACAC}2.689 & \cellcolor[HTML]{FFDB9B}3.789 & 4.526 \\
    \toprule
  \end{tabular}
  \end{subtable}

    \begin{subtable}{\textwidth}
      \centering
    \caption{\small{Mean Squared Error of Volume Change $\bm{\downarrow}$ on DFaust.}}
    \setlength{\tabcolsep}{0.5em}
    \begin{tabular}{l|l|c|c|c|c|c}
    \toprule
    \multicolumn{2}{c|}{} & \multicolumn{5}{c}{Rollout (\# of frames)} \\
    \multicolumn{2}{c|}{} & 2 & 4 & 8 & 16 & 30 \\
    \hline
    \multicolumn{7}{c}{\textit{Interpolation Set}} \\
    \hline
    \multirow{4}{*}{Non-Autoregressive} & SNARF~\cite{chen2021snarf} & 0.01582 & 0.01552 & 0.01610 & 0.01658 & 0.01682 \\
    
    & Pose & 0.01355 & 0.01305 & 0.01341 & 0.01367 & 0.01387 \\
    
    & PoseTCN  & 0.01364 & 0.01323 & 0.01350 & 0.01399 & 0.01416 \\
    
    & Pose + dPose & 0.01288 & 0.01247 & 0.01273 & \cellcolor[HTML]{FFDB9B}0.01311 & \cellcolor[HTML]{FFDB9B}0.01321 \\
    \midrule
    
    \multirow{3}{*}{Autoregressive} & G-embed & 0.01179 & \cellcolor[HTML]{FFDB9B}0.01168 & \cellcolor[HTML]{FFACAC}0.01199 & \cellcolor[HTML]{FFACAC}0.01248 & \cellcolor[HTML]{FFACAC}0.01265 \\
    
    & L-embed & \cellcolor[HTML]{FFDB9B}0.01003 & 0.01180 & 0.01466 & 0.01716 & 0.01844 \\
    
    
    & Ours & \cellcolor[HTML]{FFACAC}0.00902 & \cellcolor[HTML]{FFACAC}0.01053 & \cellcolor[HTML]{FFDB9B}0.01258 & 0.01456 & 0.01565 \\
    
    \hline
    \multicolumn{7}{c}{\textit{Extrapolation Set}} \\
    \hline
    \multirow{4}{*}{Non-Autoregressive} & SNARF~\cite{chen2021snarf} & 0.01178 & 0.01194 & 0.01251 & 0.01228 & 0.01206 \\
    
    & Pose & 0.01027 & 0.01039 & 0.01074 & 0.01052 & 0.01039 \\
    
    & PoseTCN & 0.01020 & 0.01038 & 0.01064 & 0.01040 & 0.01029 \\
    
    & Pose + dPose & 0.00992 & 0.01014 & 0.01048 & \cellcolor[HTML]{FFDB9B}0.01029 & \cellcolor[HTML]{FFDB9B}0.01013 \\
    \midrule
    
    \multirow{3}{*}{Autoregressive} & G-embed & 0.00936 & 0.00959 & \cellcolor[HTML]{FFDB9B}0.00995 & \cellcolor[HTML]{FFACAC}0.00996 & \cellcolor[HTML]{FFACAC}0.00998 \\
    
    & L-embed & \cellcolor[HTML]{FFDB9B}0.00648 & \cellcolor[HTML]{FFDB9B}0.00821 & 0.01100 & 0.01308 & 0.01402 \\
    
    
    & Ours & \cellcolor[HTML]{FFACAC}0.00567 & \cellcolor[HTML]{FFACAC}0.00715 & \cellcolor[HTML]{FFACAC}0.00915 & 0.01039 & 0.01107 \\
    
    \toprule
  \end{tabular}
\end{subtable}
\end{table}

%% file: fig/vs_nonauto.tex
\begin{figure*}[!t]
	\centering
    \includegraphics[width=0.90\linewidth]{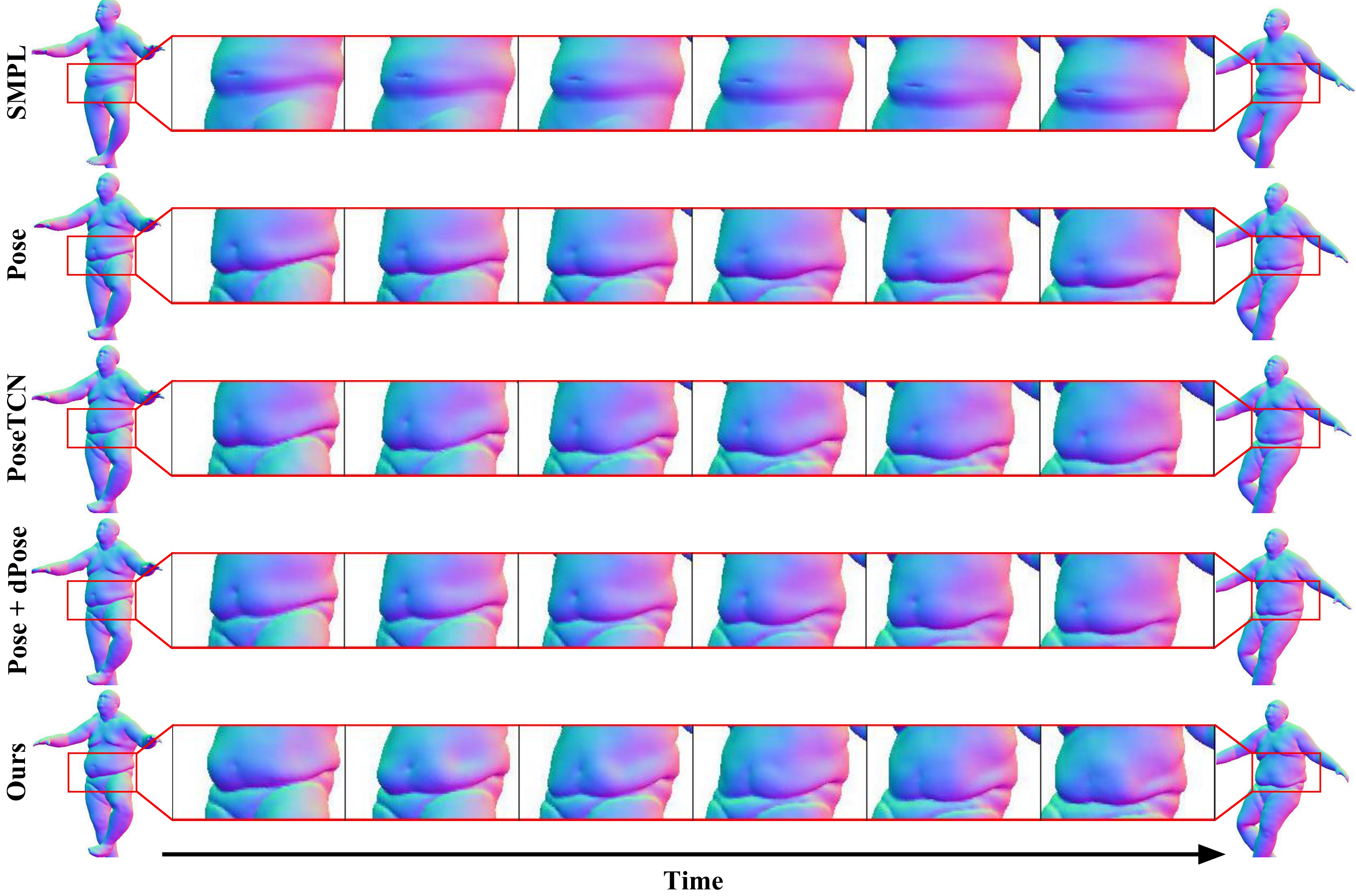}
    \caption{\small{\textbf{Qualitative Comparison with Non-Autoregressive Baselines.} In contrast to the rigid results in non-autoregressive baselines, our approach produces high quality non-rigid dynamics.}}
	\label{fig:non_autoregr}
    \vspace{-12pt}
\end{figure*}

%% file: fig/vs_latent.tex
\begin{figure*}[!t]
	\centering
    \includegraphics[width=0.90\linewidth]{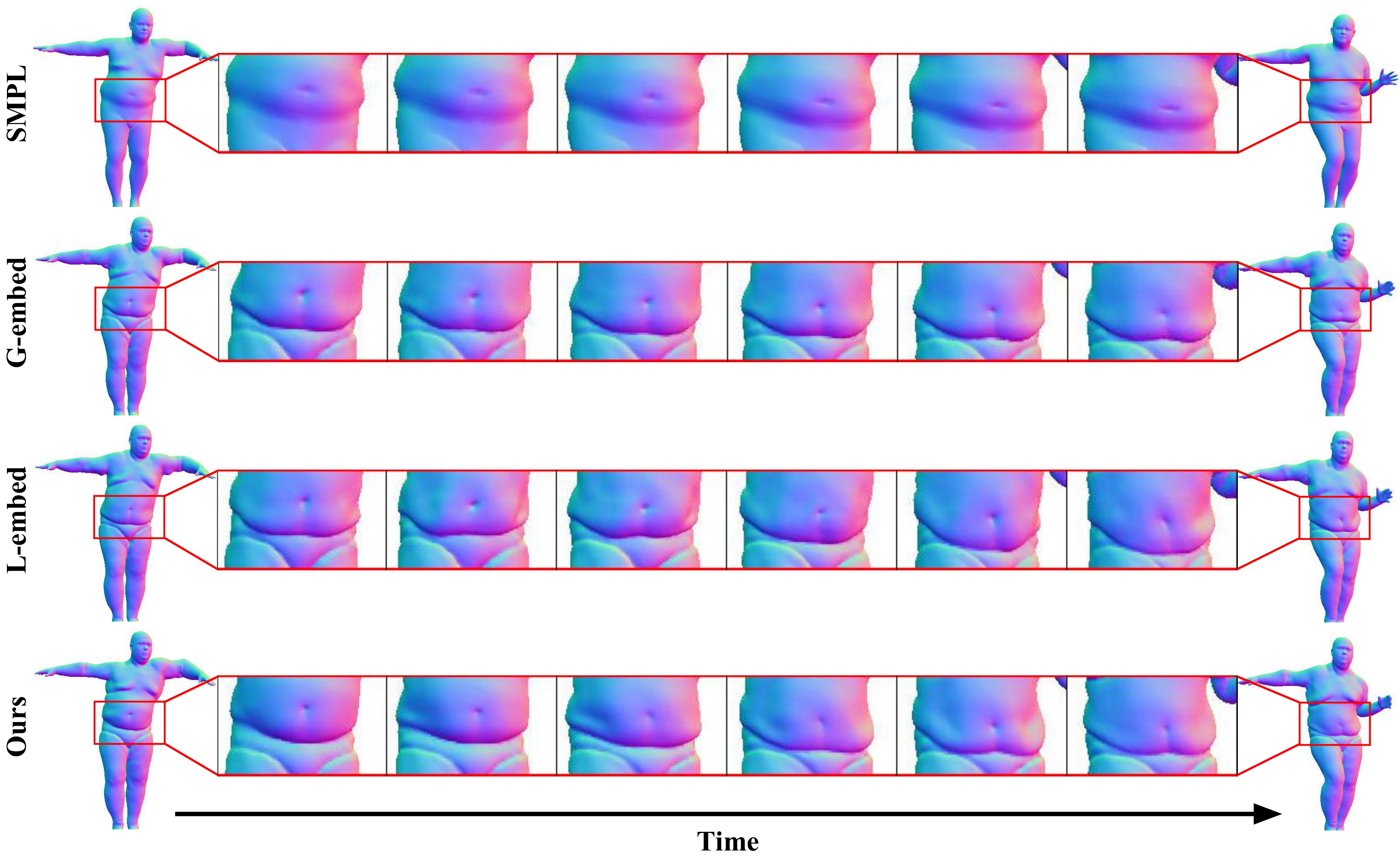}
    \caption{\small{\textbf{Qualitative Comparison with Latent-space Autoregression.} While the latent space-based autoregression approaches suffer either from overfitting to pose parameters (G-embed) or instability (L-embed, see Supp. Mat. video), our approach based on a physically meaningful quantity (signed height) achieves the most stable and expressive synthesis of dynamics.}}
    \label{fig:latent_feat_autoregr}
    \vspace{-12pt}
\end{figure*}

%% file: fig/vs_sota_vis.tex
\begin{figure*}[!t]
	\centering
    \includegraphics[width=0.8\linewidth]{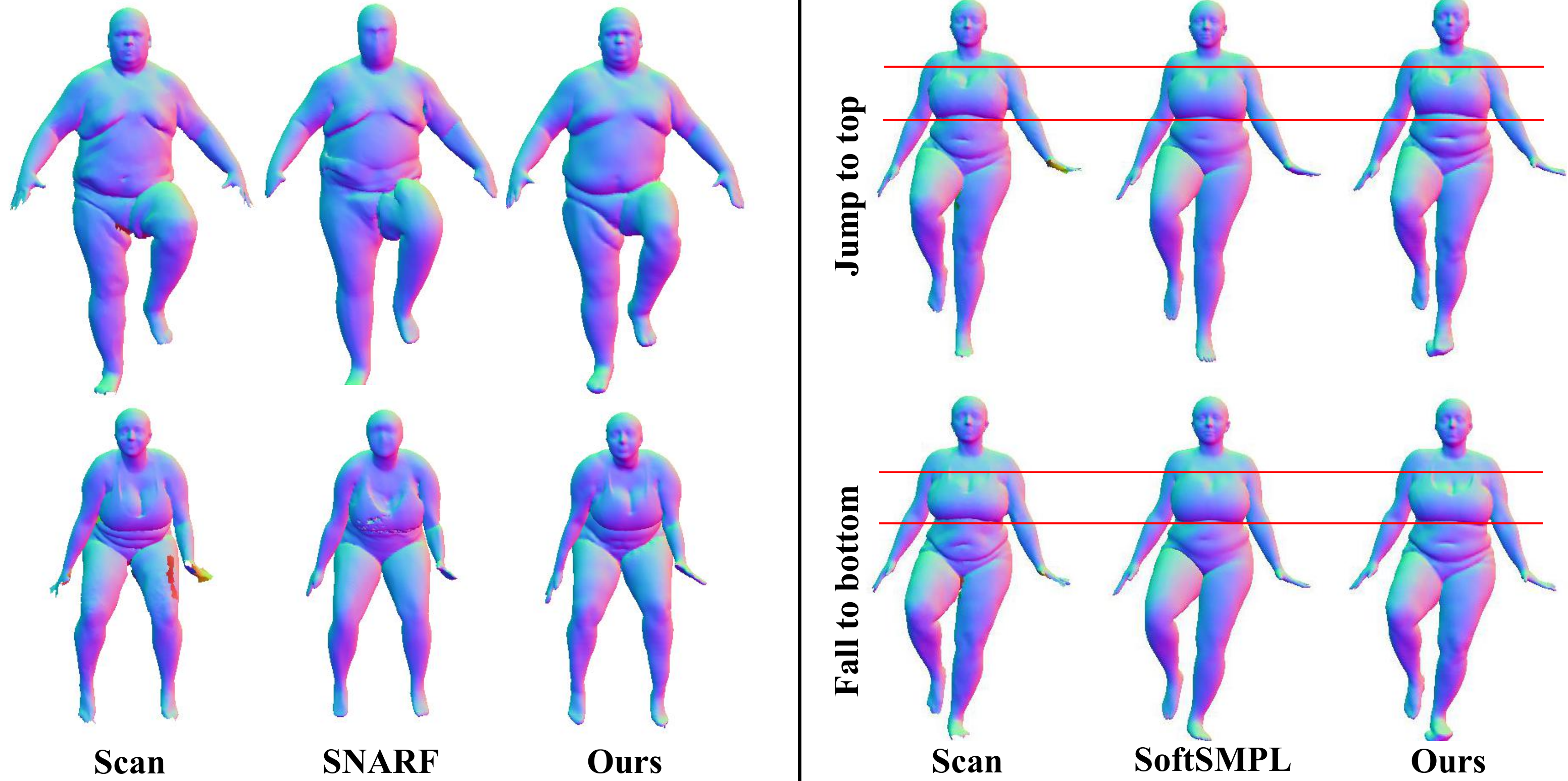}
    \caption{\small{\textbf{Qualitative Comparison with SoTA Methods.} Our approach produces significantly more faithful shapes and dynamics than the state-of-the-art implicit avatar modeling method~\cite{chen2021snarf}, and shows comparable dynamics with prior art dependent on registrations with fixed topology~\cite{santesteban2020softsmpl}.}}
	\label{fig:prior_works}
    \vspace{-8pt}
\end{figure*}

%% file: fig/vs_sota_plot.tex
\begin{figure}[t]
	\centering
    \begin{subfigure}{.5\textwidth}
      \centering
      \includegraphics[width=0.8\linewidth]{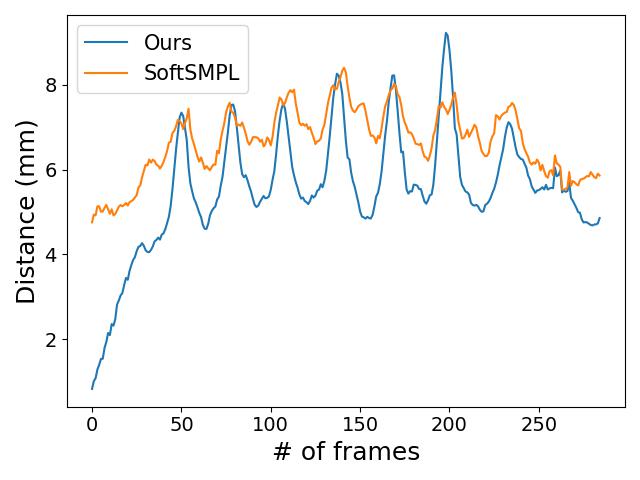}
      \caption{Scan-to-Prediction Distance $\bm{\downarrow}$}
      \label{fig:vs_softsmpl_sd}
    \end{subfigure}%
    \begin{subfigure}{.5\textwidth}
      \centering
      \includegraphics[width=0.8\linewidth]{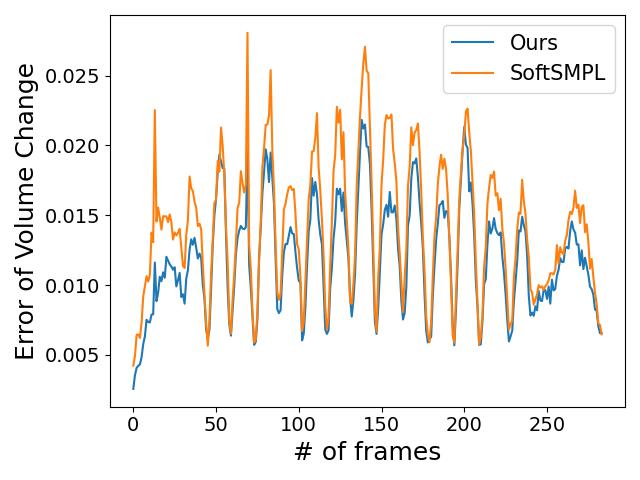}
       \caption{MSE of Volume Change $\bm{\downarrow}$}
      \label{fig:vs_softsmpl_vol}
    \end{subfigure}
    \caption{\small{\textbf{Comparison with SoftSMPL~\cite{santesteban2020softsmpl}.} We plot the errors on the sequence of \texttt{one\_leg\_loose} for subject 50004. Surprisingly, our registration-free approach mostly outperforms this baseline
    that has to rely on registered data with fixed topology.}}
	\label{fig:vs_softsmpl}
    \vspace{-16pt}
\end{figure}

%% file: sections/conclusion.tex
\section{Conclusion}
\label{sec:conclusion}
\vspace{-0.2cm}

We have introduced AutoAvatar, an autoregressive approach for modeling high-fidelity dynamic deformations of human bodies directly from raw 4D scans using neural implicit surfaces.
The reconstructed avatars can be driven by pose parameters, and automatically incorporate secondary dynamic effects that depend on the history of shapes.
Our experiments indicate that modeling dynamic avatars without relying on accurate registrations is made possible by choosing an efficient representation for our autoregressive model.

\noindent\textbf{Limitations and Future Work.}
While our method has shown to be effective in modeling the elastic deformations of real humans, we observe that it remains challenging, yet promising, to model clothing deformations that involve high-frequency wrinkles (see \cref{sec:limitations} for details).
Our evaluation also suggests that ground-truth comparison with longer rollouts may not reliably reflect the plausibility of dynamics. Quantitative metrics that handle the high sensitivity to initial conditions in dynamics could be further investigated.
Currently, AutoAvatar models subject-specific dynamic human bodies, but generalizing it to multiple identities, as demonstrated in registration-based shape modeling~\cite{pons2015dyna,loper2015smpl,santesteban2020softsmpl}, is an interesting direction for future work.
The most exciting venue for future work is to extend the notion of dynamics to image-based avatars~\cite{peng2021neural,liu2021neural}.
In contrast to implicit surfaces, neural radiance fields~\cite{mildenhall2020nerf} do not have an explicit ``surface" as they model geometry using density fields.
While this remains an open question, we believe that our contributions in this work such as efficiently modeling the state of shapes via articulated observer points might be useful to unlock this application.

%% file: supplemental/supp_content.tex
\renewcommand\thesection{\Alph{section}}
\renewcommand\thetable{\Alph{table}}
\renewcommand\thefigure{\Alph{figure}}


\section{Analysis on Input Pose Accuracy}
\label{sec:better_poses}

We investigate how the accuracy of the input SMPL fitting influences the results on subject 50002 of DFaust~\cite{bogo2017dynamic}.
As discussed in \cref{sec:datasets_metrics}, the fitted SMPL parameters in DFaust are provided by the AMASS~\cite{mahmood2019amass} dataset that uses sparse points on the registered data as approximated motion capture marker locations and computes the parameters using MoSh~\cite{loper2014mosh}. 
We observe that the provided pose parameters sometimes exhibit small misalignment with respect to the input scans.
While the fitting quality in the AMASS dataset is sufficient for our approach, we also evaluate the performance on more accurate pose parameters by using all the vertices on the registered meshes. 
More specifically, we first compute a better template by unposing the registered meshes in the first frame of each sequence using the LBS skinning weights of the SMPL template, 
and averaging over all the sequences.
Using this new template, we optimize pose parameters for each frame with an L2-loss on all the registered vertices.
Note that in this experiment, we use the original template with the refined pose parameters instead of the refined template in order not to unfairly favor our method over SNARF~\cite{chen2021snarf}.

In \cref{tab:quant_eval}, we report the mean absolute error of scan-to-prediction distance (mm) and the mean squared error of volume change for our method and SNARF. 
\cref{tab:quant_eval} shows that SNARF has a large error reduction with refined poses, indicating that SNARF is highly sensitive to the accuracy of the SMPL fit. 
We also observe that after pose refinement, SNARF overfits more to training poses (e.g., interpolation) as SNARF cannot model history-dependent dynamic deformations.
In contrast, our method is more robust to the fitting errors, and significantly outperforms SNARF in most settings except for $16$-$30$ rollouts in the interpolation set.
Note that the results with longer rollouts favor ``mean" predictions over more dynamic predictions, and do not inform us of the plausibility of the synthesized dynamics (see the discussion in \cref{sec:eval}).

\input{supplemental/quant_eval_betterposes}

\section{k-NN vs. Closest Surface Projection}
\label{sec:knn_vs_proj}

\begin{figure}
    \centering
    \includegraphics[width=\linewidth]{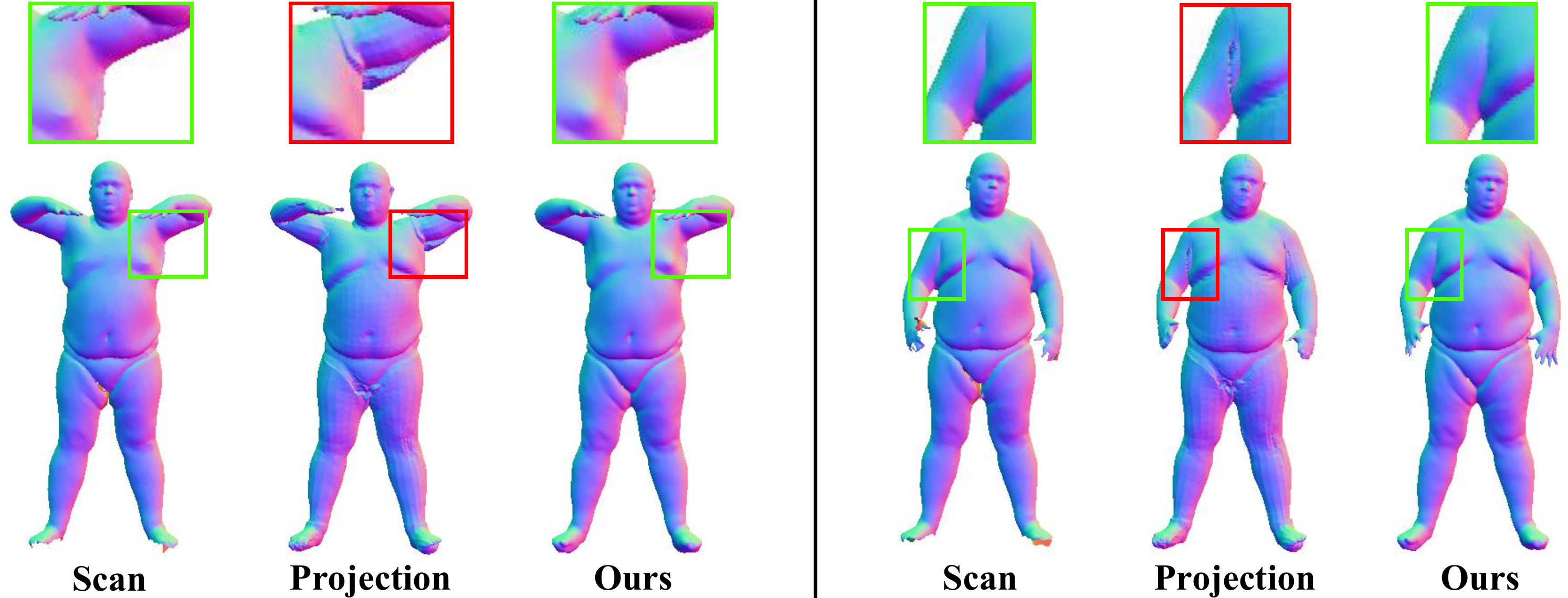}
    \caption{\small{\textbf{k-NN vs. Closest Surface Projection.} 
    While the closest surface projection suffers from artifacts around armpits, our SDF decoding based on k-NN produces more plausible surface geometry for unseen poses.
    }}
    \label{fig:knn_vs_proj}
\end{figure}

As discussed in \cref{sec:imp_dec}, our SDF decoding approach uses k-nearest neighbors (k-NN) 
of the SMPL vertices instead of closest surface projection~\cite{liu2021neural}.
\cref{fig:knn_vs_proj} illustrates the limitation of this alternative approach proposed in Neural Actor~\cite{liu2021neural}. 
As shown in \cref{fig:knn_vs_proj}, we observe that associating a query location with a single closest point on the surface leads to poor generalization to unseen poses around regions with multiple body parts in close proximity (e.g. around armpits). 
In contrast, our approach, which associates query points with multiple k-NN vertices, produces more plausible surface geometry even for unseen poses.

\section{Limitation: Clothing Deformations}
\label{sec:limitations}

We also apply our method on the CAPE~\cite{ma2020learning} dataset that contains 4D scans of clothed humans. 
We select the subject \texttt{03375\_longlong}, which exhibits the most visible dynamic deformations for clothing.
We exclude $6$ sequences (\texttt{athletics}, \texttt{frisbee}, \texttt{volleyball}, \texttt{box\_trial1}, \texttt{swim\_trial1}, \texttt{twist\_tilt\_trial1}) from training, and use them for testing. 
We employ as input the template and SMPL poses provided by the CAPE dataset for training our model.
Note that we approximate raw scans by sampling point clouds with surface normals 
computed on the registered meshes as the CAPE dataset only provides registered meshes for \texttt{03375\_longlong}.

Please refer to the supplementary video for qualitative results. 
While our approach produces plausible short-term clothing deformations, it remains challenging to model dynamically deforming clothing with longer rollouts. 
Compared to soft-tissue deformations, dynamics on clothed humans involve high-frequency deformations and topology change, making the learning of clothing dynamics more difficult. 
We leave this for future work.

%% file: supplemental/quant_eval_betterposes.tex
\begin{table}
\caption{\small{\textbf{Quantitative Evaluation on Input Pose Accuracy on Subject 50002.} 
We show the results of our approach and SNARF~\cite{chen2021snarf} using the poses provided by the AMASS~\cite{mahmood2019amass} dataset and the ones after refinement using all vertices in the registered meshes. While SNARF is greatly influenced by the accuracy of pose parameters, the slight improvement in our method illustrates its robustness to SMPL fitting errors. In addition, our approach significantly outperforms SNARF even after pose refinement in most settings except for the $16$-$30$ rollouts in the interpolation set.
}}
  \label{tab:quant_eval_better_pose}

    \begin{subtable}{\textwidth}
      
  \centering
  \caption{\small{Mean Scan-to-Prediction Distance (mm)     \label{tab:results_dfaust_sd}
$\bm{\downarrow}$}}
    \setlength{\tabcolsep}{0.5em}
    \begin{tabular}{l|l|c|c|c|c|c|c}
    \toprule
    \multicolumn{2}{c|}{} & \multicolumn{6}{c}{Rollout (\# of frames)} \\
    \multicolumn{2}{c|}{} & 1 & 2 & 4 & 8 & 16 & 30 \\
    \hline
    \multicolumn{8}{c}{\textit{Interpolation Set}} \\
    \hline

    \multirow{2}{*}{AMASS~\cite{mahmood2019amass}} & SNARF~\cite{chen2021snarf} & 7.898 & 7.715 & 7.588 & 7.840 & 7.898 & 8.238  \\
    
    & Ours & 1.731 & 2.127 & 2.953 & 4.325 & 5.606 & 6.455 \\
    \midrule
  
    \multirow{2}{*}{Refined Poses} & SNARF~\cite{chen2021snarf} & 3.982 & 4.001 & 3.964 & 4.068 & \cellcolor[HTML]{FFACAC}4.029 & \cellcolor[HTML]{FFACAC}4.158 \\
    
    & Ours & \cellcolor[HTML]{FFACAC}1.417 & \cellcolor[HTML]{FFACAC}1.703 & \cellcolor[HTML]{FFACAC}2.259 & \cellcolor[HTML]{FFACAC}3.241 & 4.044 & 4.601 \\
    
    \hline
    \multicolumn{8}{c}{\textit{Extrapolation Set}} \\
    \hline
    \multirow{2}{*}{AMASS~\cite{mahmood2019amass}} & SNARF~\cite{chen2021snarf} & 8.083 & 8.126 & 8.160 & 8.246 & 8.050 & 8.025 \\

    & Ours & 1.259 & 1.479 & 1.984 & 2.883 & 4.023 & 4.867 \\

    \midrule
    
    \multirow{2}{*}{Refined Poses} & SNARF~\cite{chen2021snarf} & 4.624 & 4.632 & 4.672 & 4.749 & 4.548 & 4.447 \\

    & Ours & \cellcolor[HTML]{FFACAC}1.149 & \cellcolor[HTML]{FFACAC}1.329 & \cellcolor[HTML]{FFACAC}1.745 & \cellcolor[HTML]{FFACAC}2.486 & \cellcolor[HTML]{FFACAC}3.313 & \cellcolor[HTML]{FFACAC}3.855 \\
    \toprule

  \end{tabular}
  \end{subtable}

    \begin{subtable}{\textwidth}
      \centering
    \caption{\small{Mean Squared Error of Volume Change $\bm{\downarrow}$}}
    \setlength{\tabcolsep}{0.5em}
    \begin{tabular}{l|l|c|c|c|c|c}
    \toprule
    \multicolumn{2}{c|}{} & \multicolumn{5}{c}{Rollout (\# of frames)} \\
    \multicolumn{2}{c|}{} & 2 & 4 & 8 & 16 & 30 \\
    \hline
    \multicolumn{7}{c}{\textit{Interpolation Set}} \\
    \hline
    \multirow{2}{*}{AMASS~\cite{mahmood2019amass}} & SNARF~\cite{chen2021snarf} & 0.01623 & 0.01590 & 0.01688 & 0.01703 & 0.01829 \\
    
    & Ours & 0.00990 & 0.01135 & 0.01417 & 0.01597 & 0.01815 \\
    \midrule
    
    \multirow{2}{*}{Refined Poses} & SNARF~\cite{chen2021snarf} & 0.01401 & 0.01349 & 0.01430 & 0.01426 & \cellcolor[HTML]{FFACAC}0.01524 \\

    & Ours & \cellcolor[HTML]{FFACAC}0.00849 & \cellcolor[HTML]{FFACAC}0.01002 & \cellcolor[HTML]{FFACAC}0.01248 & \cellcolor[HTML]{FFACAC}0.01389 & 0.01558 \\
    
    \hline
    \multicolumn{7}{c}{\textit{Extrapolation Set}} \\
    \hline
    \multirow{2}{*}{AMASS~\cite{mahmood2019amass}} & SNARF~\cite{chen2021snarf} & 0.01228 & 0.01244 & 0.01333 & 0.01292 & 0.01264 \\

    & Ours & 0.00602 & 0.00756 & 0.00977 & 0.01082 & 0.01140 \\
    \midrule
    
    \multirow{2}{*}{Refined Poses} & SNARF~\cite{chen2021snarf} & 0.01094 & 0.01092 & 0.01148 & 0.01099 & 0.01080 \\

    & Ours & \cellcolor[HTML]{FFACAC}0.00559 & \cellcolor[HTML]{FFACAC}0.00691 & \cellcolor[HTML]{FFACAC}0.00871 & \cellcolor[HTML]{FFACAC}0.00953 & \cellcolor[HTML]{FFACAC}0.01000 \\
    
    \toprule

  \end{tabular}
\end{subtable}
\end{table}